\documentclass[sigconf]{acmart}

\makeatletter
\def\mdseries@tt{m}
\makeatother 
\usepackage[draft=true,newfloat]{minted} %2
\usepackage{graphicx}
\usepackage{hyperref}           %5

\usepackage{balance}

\usepackage[utf8]{inputenc}
\usepackage{verbatim}
\newsavebox{\mintedbox}
\usepackage{multirow}
\usepackage{xcolor}
\usepackage{float}
\usepackage{algorithmic}
\usepackage[linesnumbered,lined,boxed]{algorithm2e}

\usepackage{mathtools}
\usepackage{setspace}
\usepackage{placeins}
\usepackage{tcolorbox}

\tcbuselibrary{minted,breakable,skins}
\usepackage{listings}
\usepackage{caption}
\usepackage{subfig}
\newtheorem{definition}{Definition}
\usepackage{pgfplots}

\usepgfplotslibrary{groupplots}
\newenvironment{code}{\captionsetup{type=listing}}{}
\SetupFloatingEnvironment{listing}{name=Listings}

\usepackage{xcolor,pifont}

\definecolor{cycle2}{RGB}{106, 191, 0}
\definecolor{cycle3}{RGB}{191, 0, 0}

\definecolor{amber}{rgb}{1.0, 0.75, 0.0}
\definecolor{awesome}{rgb}{1.0, 0.13, 0.32}
\definecolor{ao(english)}{rgb}{0.0, 0.5, 0.0}

\newcommand{\cmark}{\textcolor{cycle2}{\ding{52}}} %
\newcommand{\xmark}{\textcolor{cycle3}{\ding{56}}}

\newcommand{\specialcell}[2][c]{%
  \begin{tabular}[#1]{@{}c@{}}#2\end{tabular}}

\title{ChemicalX: A Deep Learning Library for Drug Pair Scoring}

\copyrightyear{2022}
\acmYear{2022}
\setcopyright{acmlicensed}\acmConference[KDD '22]{Proceedings of the 28th ACM SIGKDD Conference on Knowledge Discovery and Data Mining}{August 14--18, 2022}{Washington, DC, USA}
\acmBooktitle{Proceedings of the 28th ACM SIGKDD Conference on Knowledge Discovery and Data Mining (KDD '22), August 14--18, 2022, Washington, DC, USA}
\acmPrice{15.00}
\acmDOI{10.1145/3534678.3539023}
\acmISBN{978-1-4503-9385-0/22/08}

\begin{document}

\author{Benedek Rozemberczki}
\affiliation{
  \institution{AstraZeneca}
  \country{United Kingdom}
}
%\email{benedek.rozemberczki@astrazeneca.com}

\author{Charles Tapley Hoyt}
\authornote{Supported by the DARPA Young Faculty Award W911NF2010255}
\affiliation{
  \institution{Harvard Medical School}
  \country{United States of America}
}
%\email{cthoyt@gmail.com}

\author{Anna Gogleva}
\affiliation{
  \institution{AstraZeneca}
  \country{United Kingdom}
}

\author{Piotr Grabowski}
\affiliation{
  \institution{AstraZeneca}
  \country{United Kingdom}
}

\author{Klas Karis}
\authornotemark[1]
\affiliation{
  \institution{Harvard Medical School}
  \country{United States of America}
}
%\email{klas_karis@hms.harvard.edu}

\author{Andrej Lamov}
\affiliation{
  \institution{AstraZeneca}
  \country{Sweden}
}

\author{Andriy Nikolov}
\affiliation{
  \institution{AstraZeneca}
  \country{United Kingdom}
}

\author{Sebastian Nilsson}
\affiliation{
  \institution{AstraZeneca}
  \country{Sweden}
}

\author{Michael Ughetto}
\affiliation{
  \institution{AstraZeneca}
  \country{Sweden}
}

\author{Yu Wang}
\affiliation{
  \institution{Vanderbilt University}
  \country{United States of America}
}

\author{Tyler Derr}
\affiliation{
  \institution{Vanderbilt University}
  \country{United States of America}
}

\author{Benjamin M. Gyori}
\authornotemark[1]
\affiliation{
  \institution{Harvard Medical School}
  \country{United States of America}
}

\renewcommand{\shortauthors}{Rozemberczki et al.}
\begin{abstract}
In this paper, we introduce \textit{ChemicalX}, a PyTorch-based deep learning library designed for providing a range of state of the art models to solve the drug pair scoring task.\footnote{The open source library is available at \url{https://github.com/astrazeneca/chemicalx}.}
The primary objective of the library is to make deep drug pair scoring models accessible to machine learning researchers and practitioners in a streamlined framework.
The design of \textit{ChemicalX} reuses existing high level model training utilities, geometric deep learning, and deep chemistry layers from the PyTorch ecosystem.
Our system provides neural network layers, custom pair scoring architectures, data loaders, and batch iterators for end users.
We showcase these features with example code snippets and case studies to highlight the characteristics of \textit{ChemicalX}.
A range of experiments on real world drug-drug interaction, polypharmacy side effect, and combination synergy prediction tasks demonstrate that the models available in \textit{ChemicalX} are effective at solving the pair scoring task.
Finally, we show that \textit{ChemicalX} could be used to train and score machine learning models on large drug pair datasets with hundreds of thousands of compounds on commodity hardware.
\end{abstract}




\maketitle

\section{Introduction}\label{sec:introduction}
\begin{figure}[h!]
\centering 
\includegraphics[scale=0.17]{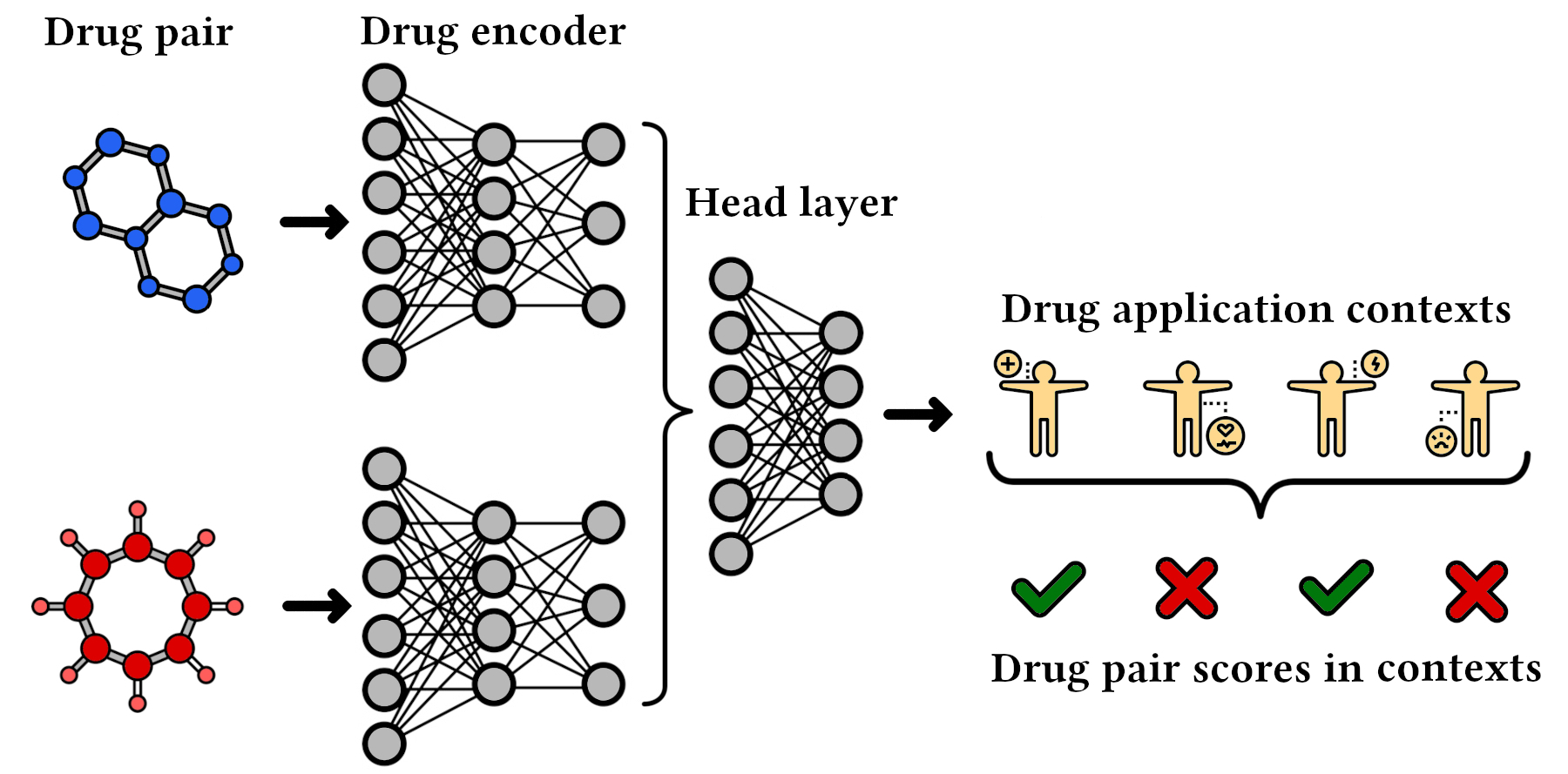}
\caption{The proposed deep learning library \textit{ChemicalX} provides data loaders, layer definitions and architectures for solving the drug pair scoring task. Models which solve this task take a drug pair as input. The drug encoder layer first learns drug representations which are combined to be drug pair representations. Using these pair representations the head layer outputs application context specific scores. }\label{fig:eyecandy}
\end{figure}

Relational deep learning has found many high-impact applications in the pharmaceutical industry in the last few years. Graph neural networks which 
operate on biological and chemical data have directly affected drug repurposing, novel target identification, and lead optimization efforts of numerous companies \cite{gaudelet2021utilizing}. This widespread adoption of relational machine learning technologies in drug discovery was made feasible thanks to the availability of domain-specific and generic open-source deep learning software \cite{torchdrug, chainerchem, fey2019fast}. These libraries are targeted at solving single drug tasks, making predictions for a single molecule which limits their applicability, because in several drug discovery scenarios--such as synergy prediction--the properties of drug pair combinations are the target of interest.
% Ben: I think we should include examples of what some of the important task are for which drug-pair scoring is necessary - I added drug synergy as an example.
The general design of deep architectures that solve the drug pair scoring task is summarized in Figure \ref{fig:eyecandy}. Our goal is to provide a unified deep learning framework to solve this task.

\textbf{Present work.} We release \textit{ChemicalX}, a PyTorch \cite{paszke2019pytorch}-based open-source deep-learning library for providing a framework that can efficiently and accurately solve the \textit{drug pair scoring task} \cite{rozemberczki2021unified}. Our library was developed by reusing basic functionalities of existing geometric deep learning and deep chemistry frameworks of the PyTorch ecosystem \cite{torchdrug, fey2019fast}. It is designed to provide data loaders, integrated benchmark datasets, drug pair scoring-specific deep learning architectures, training, and evaluation routines. Models in our framework can be used standalone or trained in a memory-efficient manner with the use of \textit{ChemicalX}-specific custom data structures. The release of the framework comes with detailed documentation, integrated benchmark datasets, and case studies in the form of example scripts and tutorials.

The empirical evaluation of \textit{ChemicalX} focused on solving traditional drug pair scoring tasks on publicly available real-world chemical datasets. We assessed and contrasted the predictive performance of deep learning models available in the library on polypharmacy side effects, drug-drug interaction, and synergy prediction tasks.  Additional experiments on real-world data demonstrate that \textit{ChemicalX} can be scaled up with commodity hardware to train on datasets that could contain millions of labeled drug pairs.

\textbf{Our contributions.} The main contributions of our work can be summarized as:
\begin{itemize}
    \item We publicly released \textit{ChemicalX}, the first deep learning library dedicated to solving the drug pair scoring task.
    % Ben: clarification - is the above true? Don't the cited models also "solve" the drug pair scoring task in some way? The difference I guess is more that ChemicalX integrates multiple models in a unified framework.
    \item We provided drug pair scoring models, drug pair dataset loaders, and iterators with \textit{ChemicalX} to facilitate data mining and cheminformatics research in the domain.
    \item We integrated public drug-drug interaction, synergy, and polypharmacy side effect prediction datasets in \textit{ChemicalX}.
    \item We evaluated the drug pair scoring models implemented in \textit{ChemicalX} on various drug pair scoring tasks and investigated the scalability of the proposed framework. 
\end{itemize}

The remainder of our work has the following structure; in Section \ref{sec:related_work} we provide an overview of related literature on molecular fingerprints, deep learning and drug pair scoring. Section \ref{sec:preliminaries} introduces core concepts about drug pair scoring and we discuss the architecture of \textit{ChemicalX} in Section \ref{sec:design}. We evaluate the drug pair scoring models implemented in \textit{ChemicalX} in Section \ref{sec:experiments}. Potential future directions are discussed in Section \ref{sec:future} and the paper concludes with Section \ref{sec:conclusions}.

\section{Related work }\label{sec:related_work}

Our work intersects with the theory of traditional molecular and deep learning-based representations. First, we discuss molecular serializations, then look at how molecular fingerprints and graphs are extracted from these. We discuss how geometric deep learning models learn from the extracted molecular graphs, overview techniques for drug pair scoring, and open-source software for deep drug discovery.

\subsection{Molecular Representation and Featurization}

Medicinal and synthetic chemists often focus on molecules' two-dimensional structures as a proxy for their three-dimensional geometries and electrostatic properties that endow their biological activities.
While they typically communicate two-dimensional structure via Kekulé diagrams, the technical limitations underlying most chemical information systems motivated the development of linear, string-based representations including SMILES~\cite{weininger1988smiles} and InChI~\cite{heller2015inchi}.

Two-dimensional structures have been classically used to generate discrete features for molecules based on the presence or absence of an enumerated list of substructures~\cite{durant2002maacs} then later based on a hash over arbitrarily generated substructures~\cite{morgan1965fingerprint}.
More recently, language models have been applied to molecular strings to learned continuous features~\cite{gomez-bombarelli2018}, which in turn motivated the development of more robust string-based representations including DeepSMILES~\cite{oboyle2018deepsmiles} and SELFIES~\cite{krenn2020selfies}.
Finally, modern techniques use graph convolutions to directly learn continuous representations of the molecular graph~\cite{mercado2021graphreinvent}.

\subsection{Graph Representation Learning}
Graph neural network layers can create expressive learned drug representations. This happens by neural message passing on the molecular graph of the drug \cite{gilmer2017neural}; atoms are treated as nodes and edges are atomic bonds on which learned neural atom representations are propagated \cite{kipf2017semi, graph2018velickovic,duvenaud2015convolutional}. The atom representations are aggregated by permutation invariant pooling functions \cite{gaudelet2021utilizing} to generate the drug representation. Several modern pair scoring architectures in \textit{ChemicalX}  use drug representations learned from molecular graphs to do pair scoring \cite{chen2020gcnbmp, xu2019mr, sun2020structure}. These models are differentiated by the message-passing layer used to generate atom representations and the pooling function that distills the drug representations.

\begin{table}[h!]
\caption{The drug pair scoring models implemented in \textit{ChemicalX} sorted by publication year with the respective application domain (polypharmacy/interaction/synergy) and the architecture of the drug encoder. }\label{tab:drug_pair_scoring}
\begin{tabular}{lccc}
\hline
\textbf{Model}  &\textbf{Year} &\textbf{Domain}  &  \textbf{Encoder}  \\
 \hline

%\textbf{DeepCCI}      \cite{kwon2017deepcci}& 2017 &Interaction  &  Feedforward  \\[0.1cm]
\textbf{DeepDDI}      \cite{ryu2018deep}&2018& Interaction  &    Feedforward \\[0.1cm]
\textbf{DeepSynergy}  \cite{preuer2018deepsynergy}&2018 &Synergy &   Feedforward \\[0.1cm]
\textbf{MHCADDI}      \cite{deac2019mhcaddi}& 2019&Polypharmacy & GAT  \\[0.1cm]
\textbf{MR-GNN}        \cite{xu2019mr}& 2019&Interaction   &    GCN \\[0.1cm]
\textbf{CASTER}       \cite{huang2019caster}& 2019&Interaction  &  Feedforward \\[0.1cm]
\textbf{SSI-DDI}      \cite{nyamabossi20201ssi}& 2020&Interaction  &  GAT\\[0.1cm]
\textbf{EPGCN-DS}     \cite{sun2020structure}&2020& Interaction  &  GCN\\[0.1cm]
%\textbf{AuDNNSynergy} \cite{zhang2021synergistic}& 2020 & Synergy & Feedforward   \\[0.1cm]
\textbf{DeepDrug}     \cite{cao2020deepdrug}&2020& Interaction  & GCN  \\[0.1cm]
\textbf{GCN-BMP}      \cite{chen2020gcnbmp}& 2020&Interaction  & GCN \\[0.1cm]
%\textbf{DPDDI}        \cite{wang2021deepdds}& 2021&Interaction  &  GCN \\[0.1cm]
\textbf{DeepDDS}     \cite{wang2021deepdds}& 2021 &Synergy & GCN or GAT \\[0.1cm]
\textbf{MatchMaker}   \cite{brahim2021matchmaker}& 2021 & Synergy &   Feedforward\\[0.1cm]
 \hline
\end{tabular}
\end{table}

\subsection{Deep Learning for Drug Pair Scoring}
Computational drug pair scoring requires learning a function that can predict scores for pairs of drugs in a biological or chemical context \cite{rozemberczki2021unified}.
In the deep learning setting, this learned function takes the form of a neural network.
Pair scoring models have three main application domains; in each of these domains, the models output the probability of a positive answer to a domain- and context-specific question about a drug pair.
These domains and questions are: 

\begin{itemize}
    \item \textbf{Polypharmacy side effects}. Is it possible that drugs \textit{X} and \textit{Y} together cause polypharmacy side effect \textit{Z}? 
    \item \textbf{Drug-drug interactions.} Can drugs \textit{X} and \textit{Y} have interaction \textit{Z} when administered together?
    \item \textbf{Pair synergy identification.} Are drugs \textit{X} and \textit{Y} synergistic at treating disease \textit{Z} when applied in combination?
\end{itemize}
% Ben: In the above, there are 3 different phrasings for saying that the two drugs are applied in combination ("together cause polyphramacy...", "administered together", "applied in combination") - we could standardize the phrasings to make it more clear.
In Table 
\ref{tab:drug_pair_scoring} we listed all of the drug pair scoring models implemented in \textit{ChemicalX} with the application domain and architectural details of the models. The prevalence of graph neural network-based encoders is evident in the newer pair scoring architectures.

\subsection{Software for Deep Drug Discovery}

Deep learning libraries designed for machine learning on chemical data are all built on open-source automatic differentiation frameworks such as TensorFlow~\cite{abadi2016tensorflow}, PyTorch~\cite{paszke2019pytorch}, MXNet~\cite{mxnet}, JAX~\cite{jax}, and Chainer \cite{tokui2015chainer}. We listed those publicly available machine learning libraries that can be used to define architectures that can solve deep drug discovery tasks in Table~\ref{tab:systems}.

\begin{table}[h!]
\caption{A comparison of deep learning libraries that operate on drugs. Libraries are ordered by year of release. We included information about automatic differentiation backend (TensorFlow - TF, MXNet - MX, PyTorch - PT, JAX, Chainer - CH), application domain and suitability for pair scoring.}\label{tab:systems}
\begin{tabular}{lcccc}
\hline
\textbf{Library} &\textbf{Year}& \textbf{Backend} & \specialcell{\textbf{Drug}\\\textbf{Domain}}&  \specialcell{\textbf{Pair}\\\textbf{Scoring}}\\
 \hline
\textbf{PyG} \cite{fey2019fast}&2018&  PT &   \xmark&\xmark  \\[0.1cm]
\textbf{DGL} \cite{wang2019deep}&2019&  PT/TF/MX &   \xmark &\xmark\\[0.1cm]
\textbf{StellarGraph} \cite{stellargraph}&2019&  TF & \xmark &\xmark\\[0.1cm]
\textbf{DeepChem} \cite{deepchem}&2019&  TF &  \cmark
& \xmark\\[0.1cm]
\textbf{CHChem} \cite{chainerchem}& 2019& CH &  \cmark& \xmark\\[0.1cm]
\textbf{Jraph}  \cite{jraph2020github}&2020& JAX              &  \xmark&\xmark\\[0.1cm]
\textbf{Spektral} \cite{spektral} &2020&  TF  & \xmark& \xmark \\[0.1cm]
\textbf{DIG} \cite{dig}& 2021 & PT & \xmark& \xmark \\[0.1cm]
\textbf{TorchDrug} \cite{torchdrug} &2021& PT& \cmark& \xmark\\[0.1cm]
\textbf{CogDL}   \cite{cen2021cogdl} &2021&  PT     &    \xmark&\xmark \\[0.1cm]
\textbf{TFG}  \cite{hu2021efficient} &2021&TF & \xmark& \xmark \\[0.1cm]
\textbf{DGL-LS}  \cite{li2021dgl}&2021& PT              &\cmark&\xmark\\[0.1cm]
\hline 
\textbf{Our Work} &  2022 & PT& \cmark& \cmark\\[0.1cm]
 \hline
\end{tabular}
\end{table}

Based on Table~\ref{tab:systems}, we can conclude that PyTorch is the most widely used automatic differentiation backend of machine learning libraries which can be used for drug discovery and deep chemistry. This is mainly due to the well-developed geometric deep learning ecosystem of this framework \cite{fey2019fast, rozemberczki2021pytorch}. Only a few of the existing libraries were designed specifically to operate on chemical data \cite{deepchem,chainerchem,torchdrug,li2021dgl}, but none of those is dedicated to drug pair scoring.

\section{Preliminaries}\label{sec:preliminaries}
Our discussion of drug pair scoring deep learning models is based on the \textit{unified view} of drug pair scoring described by \citep{rozemberczki2021unified}. To facilitate the description of \textit{ChemicalX} features and architecture design, we introduce formal definitions related to the drug pair scoring task.

\subsection{The unified drug pair scoring model}
We assume that we have a set of drugs $\mathcal{D}=\left\{d_1,\dots,d_n\right\}$ for which we know the chemical structure of molecules and a set of classes $\mathcal{C}=\left\{c_1,\cdots,c_p\right\}$ that describes the types of contexts in which a drug pair can be administered.

\begin{definition}\label{def:drugset} \textbf{Drug feature set.} Given drug set $\mathcal{D}$ a drug feature set is the set of tuples $(\textbf{x}^{d}, \mathcal{G}^{d},\textbf{X}^{d}_N,\textbf{X}^{d}_E)\in \mathcal{X}_{\mathcal{D}},\,\,\forall d \in \mathcal{D}$, where $\textbf{x}^d$ is the molecular feature vector, $\mathcal{G}^d$ is the molecular graph of the drug, $\textbf{X}^{d}_N$ and $\textbf{X}^{d}_E$ are the atom and the edge feature matrices.
\end{definition}
We assume that drugs can be described with 4 types of information; (i) Molecular features which give high-level information about the molecule such as measures of charge. (ii) The molecular graph in which nodes are atoms and edges are bonds. (iii) Node features in the molecular graph such as the type of the atom or whether it participates in an aromatic ring. (iv) Edge features that describe the bonds in the molecular graph such as the type of the bond.

\begin{definition}\label{def:contextset} \textbf{Context feature set.} Given context set $\mathcal{C}$ a context feature set is the set of context feature vectors $\textbf{x}^{c}\in \mathcal{X},\,\,\forall c \in \mathcal{C}$.
\end{definition}
A context feature set allows for making context-specific predictions that take into account the similarity of the contexts. For example in a synergy prediction setting the context features can describe the gene expressions in the targeted cancer cell.
\begin{definition} \textbf{A labeled drug pair - context triple set.}\label{def:labeled_triples} A labeled drug pair - context set is the set of tuples  $(d,d',c,y^{d,d',c})\in\mathcal{Y}$ where $d,d'\in \mathcal{D}$, $c\in \mathcal{C}$ and $y^{d,d',c}\in \{0,1\}$.
\end{definition}

A set of labels contains binary target variables for a drug pair in a specific biological or chemical context. In a synergy prediction setting, this could describe whether a drug pair is synergistic at killing a specific type of cancer cell. If the labels describe a continuous outcome the definition can be modified to include that $y^{d,d',c}\in \mathbb{R}$.
\begin{definition} \textbf{Drug encoder.} This encoder defined by Equation \eqref{eq:drug_encoder} is a multivariate parametric function $f_{\Theta_D}(\cdot)$ parametrized by $\Theta_D$ which depends on the molecular feature vector $\textbf{x}^d$, molecular graph $\mathcal{G}^d$, atom and edge feature matrices $\textbf{X}^{d}_N$ and $\textbf{X}^{d}_E$. For each drug in the drug set it outputs $\textbf{h}^d$ a vector representation of the drug.
\begin{align}
\textbf{h}^{d}&=f_{\Theta_D}(\textbf{x}^{d},\mathcal{G}^{d},\textbf{X}^{d}_N,\textbf{X}^{d}_E),\,\,\,\,\forall d\in \mathcal{D}\label{eq:drug_encoder}
\end{align}
\end{definition}
A drug encoder is a neural network which maps the molecular features into a low dimensional vector space. It can use various architectures such as feedforward neural networks or graph neural networks to achieve this.
\begin{definition} \textbf{Context encoder.} The context encoder defined by Equation \eqref{eq:context_encoder} is the parametric function $f_{\Theta_C}(\cdot)$ parametrized by $\Theta_C$ which depends on the context feature vector $\textbf{x}^c$.
\begin{align}
\textbf{h}^{c}&=f_{\Theta_C}(\textbf{x}^{c}),\,\,\,\,\forall c\in \mathcal{C}\label{eq:context_encoder}
\end{align}
\end{definition}

A context encoder is a neural network that outputs a low dimensional representation of the biological context; this serves as intermediate input for the drug pair scoring decisions.

\begin{definition} \textbf{Scoring head layer.} The scoring head layer described by Equation \eqref{eq:scoring_head} is the parametric function $f_{\Theta_H}(\cdot)$ parametrized by $\Theta_H$ which depends on the drug representations $\textbf{h}^d$ and $\textbf{h}^{d'}$ for the drug pair $d,d' \in \mathcal{D}$ and the context representation $\textbf{h}_c$ for the context $c\in \mathcal{C}$. It outputs $\widehat{y}^{d,d',c}$ the estimated probability of a positive label for the drug pair - context triple.
\begin{align}
    \widehat{y}^{d,d',c} = f_{\Theta_H}(\textbf{h}^{d}, \textbf{h}^{d'}, \textbf{h}^c),\,\,\,\,\forall d,d' \in \mathcal{D},\forall c \in \mathcal{C}\label{eq:scoring_head}
\end{align}
\end{definition}

The scoring head is the final neural network layer -- it depends on the drug and biological context representations output by the respective encoders. It outputs the probability for a positive label that is used for the computation of loss values. 

\begin{definition} \textbf{Drug pair scoring cost and loss.} The drug pair scoring cost defined on labeled drug pair set $\mathcal{Y}$ in Equation \eqref{eq:cost} is the sum of loss function values. Drug pair loss values depend on the ground truth  ${y}^{d,d',c}$ and predicted labels $\widehat{y}^{d,d',c}$.
\begin{align}
\mathcal{L}=\smashoperator[r]{ \sum _{(d,d',c,y^{d,d',c})\in \mathcal{Y}}}\overbrace{\ell(y^{d,d',c},\widehat{y}^{d,d',c})}^{\text{Loss on the triple d,d',c.}} \label{eq:cost}
\end{align}
\end{definition}

The drug pair scoring cost in Equation \eqref{eq:cost} is a function of the encoder and the scoring head layer parameters defined by Equations \eqref{eq:drug_encoder}, \eqref{eq:context_encoder} and \eqref{eq:scoring_head}. Our goal is to minimize this cost by finding the optimal parametrization ($\Theta_D, \Theta_C$, and $\Theta_H$) of the layers. 
\subsection{Training a drug pair scoring model}
A drug pair scoring model can be trained by backpropagation and an appropriate variant of gradient descent which minimizes the cost described by Equation \eqref{eq:cost}. The computation of this cost requires a forward propagation step described in Algorithm \ref{alg:training}. Our design of \textit{ChemicalX} assumes a batched forward propagation flow as described in this algorithm.
	\begin{algorithm}[h!]
{\small		\DontPrintSemicolon
		\SetAlgoLined
		\KwData{
		$\mathcal{X}_D$ - Drug feature set.\\
		$\quad\quad\,\,\,\,\mathcal{X}_C$ - Context feature set.\\
		$\quad\quad\,\,\,\,\mathcal{Y}$ - Labeled drug pair - context triple set.}
		\KwResult{$\mathcal{L}$ - The cost for the labeled drug pair - context set.}

$\mathcal{L}\leftarrow 0 $\;
\For{$(d,d',c,y^{d,d',c}) \in \mathcal{Y}$}{
$\textbf{h}^{d}\leftarrow f_{\Theta_D}(\textbf{x}^{d}, \mathcal{G}^{d}, \textbf{X}^d_{N}, \textbf{X}^d_{E} )$\;
$\textbf{h}^{d'}\leftarrow f_{\Theta_D}(\textbf{x}^{d'}, \mathcal{G}^{d'}, \textbf{X}^{d'}_{N}, \textbf{X}^{d'}_{E} )$\;
$\textbf{h}^{c}\leftarrow f_{\Theta_C}(\textbf{x}^{c})$\;
$\widehat{y}^{d,d',c}\leftarrow f_{\Theta_H}(\textbf{h}^d, \textbf{h}^{d'}, \textbf{h}^{c})$\;

$\mathcal{L}\leftarrow  \mathcal{L} + \ell(y^{d,d',c}, \widehat{y}^{d,d',c})$\;
}
}
		\caption{A general cost calculation algorithm for deep learning models which can solve the drug pair scoring task.}\label{alg:training}
	\end{algorithm}

Computing the cost output in Algorithm \ref{alg:training} requires the feature sets and the labeled drug pair-context set. The cost is set to zero and the algorithm iterates over the labeled drug pair-context set (lines 1-2). For both drugs, in the drug pair, the drug encoder outputs the vector representations (lines 3-4). The context encoder creates a context representation which is used along with the drug representations by the head layer to predict the label (lines 5-6). The loss computed for a drug pair-context triple is added to the accumulated cost (line 7). After the cost for the whole labeled triple set is accumulated backpropagation happens and the weights of the respective encoders and head layer are updated.

\section{The framework design}\label{sec:design}
Our main contribution is a specialized deep learning library for solving the drug pair scoring task. We overview the data structures used in our library, discuss the design in action and highlight how we ensure that \textit{ChemicalX} is maintained in the future.

\subsection{Data Structures and Model Classes}
The design of data structures, batch generators, drug pair batch, and model classes are practical conceptualizations of the definitions and theoretical ideas outlined in Section \ref{sec:preliminaries}. 

\subsubsection{Drug and context feature sets.} Drug feature sets described by Definition \ref{def:drugset} are modeled as customized nested hashmaps. Drug identifiers are used as keys in the top-level hashmap and the drug features are values in these hashmaps. Each drug feature value is a hashmap itself with two key-value pairs. One of these key-value pairs contains the TorchDrug \cite{torchdrug} molecular graph created from canonical SMILES \cite{weininger1988smiles} representation of the drug.  This molecular graph also contains the atom and bond features of the graph. Another one is a feature vector -- the Morgan fingerprint \cite{morgan1965fingerprint} of the drug molecule stored as a PyTorch tensor.  Context feature sets conceptualized by Definition \ref{def:contextset} are implemented as customized hashmaps where \textit{keys} are application context identifiers and \textit{values} are features of the context. Individual biological and chemical context feature vectors are stored as PyTorch tensors. Using these allows the linear time retrieval of the respective features.

\subsubsection{Labeled triples, batch generators, and drug pair batches.} In Definition \ref{def:labeled_triples} we described the labeled drug pair-context sets that contain drug pairs and contexts from the respective sets with target labels. We conceptualized this as a wrapper class around Pandas dataframes; labels, drug, and context identifiers can be stored in a columnar format. The labeled triple set instances can be split into training and test set labeled drug pair-context triple instances with a class method.  We define batch generators in \textit{ChemicalX} that provide drug pair batches using the labeled drug pair - context triples, drug and context feature sets. A drug pair batch is a custom data class that holds the molecular graph, drug features, context features, and the label for each compound in a batch of drug pairs. Instances of drug pair batches are created by the generator using the labeled drug pair - context triples, the drug and context feature sets. The main advantage of this design is that the drug and context features do not have to be stored for all of the drug pairs in the main memory, but can be collated dynamically. Batched molecular graphs are returned by the generators as TorchDrug PackedGraph instances for both drugs in the drug pairs while drug and context features are returned as PyTorch FloatTensor instances.

\subsubsection{Model classes, auxiliary layers, and pipelines.} Deep drug pair scoring architectures are implemented in \textit{ChemicalX} as PyTorch neural network modules. Models are built written in pure PyTorch, except for the graph neural network drug encoder and graph pooling layers \cite{kipf2017semi,graph2018velickovic} which use the existing graph convolutional and graph attention layers of TorchDrug. This design choice allows smooth interfacing with the molecular graphs returned in the drug pair batches by the batch generator. Model hyperparameters such as the number of hidden layer channels in drug encoders have sensible default settings; this allows the out of box use of the pair scoring models by non-machine learning expert end-users. Model architectures sometimes require the definition of custom auxiliary neural network layers these are available on the same namespace as the deep pair scoring model which uses the auxiliary layer. Finally, \textit{ChemicalX} comes with a high-level end-to-end training pipeline utility function which was influenced by the design of PyKeen \cite{ali2021pykeen}.

\subsection{Design in Action - an Oncology Use Case}
We are showcasing the design of \textit{ChemicalX} by overviewing a case study from computational oncology in detail. In this task we will load the \textit{DrugComb} dataset \cite{zagidullin2019drugcomb,zheng2021drugcomb} which contains known synergistic drug pairs that are effective at destroying certain cancer cell types. Using this dataset we train and score a \textit{DeepSynergy} model \cite{preuer2018deepsynergy} which can solve the drug pair synergy scoring task. Our discussion is a tutorial accompanied by example Python scripts.

\begin{code}
\begin{minted}[linenos,fontsize=\small,xleftmargin=0.5cm,numbersep=3pt,frame=lines]{python}
from chemicalx.data import DrugCombDB, BatchGenerator

loader = DrugCombDB()

context_set = loader.get_context_features()
drug_set = loader.get_drug_features()
triples = loader.get_labeled_triples()

train, test = triples.train_test_split(train_size=0.5)

generator = BatchGenerator(batch_size=1024,
                           context_features=True,
                           drug_features=True,
                           drug_molecules=False,
                           context_feature_set=context_set,
                           drug_feature_set=drug_set,
                           labeled_triples=train)
\end{minted}
\captionof{listing}{Loading the drug, context and labeled triple sets of the DrugComb dataset. Creating a train-test split from the labels and initializing a data generator for batching.}\label{code:loading}
\end{code}
\subsubsection{Loading data and defining a generator.}
Our first step is to load the dataset in the previously discussed data structures, create a train-test split and initialize a batch generator. An example Python workflow to achieve this is laid out in Listings \ref{code:loading}. The benchmark dataset loader and the batch generator classes are imported (line 1). A dataset loader instance is initialized with parametrization that allows the loading of the DrugComb dataset (line 3). The context, drug, and labeled triple sets are all loaded in memory (lines 5-7). Using a class method the original triples are halved to create training and test triples (line 9). A batch generator is defined that will generate batches of $2^{10}$ drug pairs using the training triples, drug-, and context feature sets (lines 11-17).

\begin{code}
\begin{minted}[linenos,fontsize=\small,xleftmargin=0.5cm,numbersep=3pt,frame=lines]{python}

import torch
from chemicalx.models import DeepSynergy

model = DeepSynergy(context_channels=112,
                    drug_channels=256)

optimizer = torch.optim.Adam(model.parameters())
model.train()
loss = torch.nn.BCELoss()

for batch in generator:
    optimizer.zero_grad()
    prediction = model(batch.context_features,
                       batch.drug_features_left,
                       batch.drug_features_right)
    loss_value = loss(prediction, batch.labels)
    loss_value.backward()
    optimizer.step()
\end{minted}
\captionof{listing}{Defining a model, optimizer, loss and using the pre-loaded dataset for fitting the model on the training set.}\label{code:training}
\end{code}

\subsubsection{Model training.} As a next step we will define the deep pair scoring model and train it with the batches obtained by the generator -- this process is described in Listings \ref{code:training}. We import the base PyTorch library \cite{paszke2019pytorch} and the DeepSynergy model from \textit{ChemicalX} (lines 1-2). We define a model and set the number of context and molecular features manually  (lines 4-5). An Adam optimizer instance is defined \cite{kingma2015adam} where the model parameters are registered, the model is set to be in training mode and we initialize the binary cross-entropy loss (lines 7-9). We generate drug pair batches from the training triples (line 11). In each step we set the gradients to be zero (line 12), a forward pass is made to generate predicted scores (lines 13-15), an average binary cross-entropy loss value is computed (line 16), backpropagation happens and the weights are updated (lines 17-18). It is worth noting that the cost calculation method in Algorithm \ref{alg:training} is described by lines 13-17 in Listings \ref{code:training}.

\begin{code}
\begin{minted}[linenos,fontsize=\small,xleftmargin=0.5cm,numbersep=3pt,frame=lines]{python}
import pandas as pd

model.eval()
generator.labeled_triples = test

predictions = []
for batch in generator:
    prediction = model(batch.context_features, 
                       batch.drug_features_left,
                       batch.drug_features_right)
    prediction = prediction.detach().cpu().numpy()
    identifiers = batch.identifiers
    identifiers["prediction"] = prediction
    predictions.append(identifiers)
predictions = pd.concat(predictions)
\end{minted}
\captionof{listing}{Using the trained pair scoring model to score the triples in the test portion of the dataset.}\label{code:evaluation}
\end{code}

\subsubsection{Model scoring.} The final step is to score on the test set  -- we do this in Listings \ref{code:evaluation}. The predictions will be stored in a columnar data format, this requires the import of the pandas library (line 1). The model is set to be in evaluation mode, triples of the generator are reset with the test data and we need a list to accumulate the predictions (lines 3-5). We generate drug pair batches from the test triples to score the drugs (line 7). In each iteration the scores are predicted for the triples in the batch, predictions are added to the data in the batch and those are added to the list of predictions which are concatenated together in the end (lines 8-15).

\subsection{Maintaining and Supporting ChemicalX}
We release the source code of \textit{ChemicalX} under a permissive Apache 2.0 license, our work comes with documentation for end-users and supports the possibility of external contributions.
\subsubsection{Open Source Code and Package Indexing.} The Python source code of \textit{ChemicalX} is publicly available in a \textit{GitHub} repository under the \textit{Apache 2.0} license. This repository also provides a detailed readme, case study like example Python scripts, unit tests, installation guidelines, and references to models and datasets. It is supported by package releases on the \textit{Python Package Index} which allows the library to be installed via the \textit{pip} command from the terminal.   

\subsubsection{Documentation.} Data structure and model classes in \textit{ChemicalX} are all documented within the source code of the library. Using the raw source code and \textit{restructured text} we maintain detailed documentation of the library. Updates to the main branch of the \textit{GitHub} repository automatically trigger a build process which redeploys the \textit{Sphinx} compiled documentation on \textit{Read The Docs}. This documentation also provides introductory-level tutorials for potential users.
\subsubsection{Unit Tests, Code Coverage, and Continuous Integration.} The \textit{ChemicalX} codebase is comprehensively covered by unit tests which allow a detailed assessment of data structure and model behavior. Tests of the model classes also serve as integration tests which provide diagnostics for the whole library. Using these tests a \textit{CodeCov} based coverage report is generated which quantifies the code coverage for the library. The public \textit{ChemicalX} repository has continuous integration setup that makes sure that the library can be installed and every feature of the library is tested. Moreover, by this, we ensure that contributions to the \textit{ChemicalX} codebase follow the guidelines concerning coding conventions and formatting.

\section{Experimental Evaluation}\label{sec:experiments}
The experimental evaluation of ChemicalX mainly focuses on the predictive performance of various drug pair scoring models and the runtime of the pair models available in the library. In our experiments, we use real-world drug pair scoring datasets from various application domains of drug pair scoring such as combination therapy in oncology and polypharmacology.
\subsection{Datasets}
In \textit{ChemicalX} we integrated and created loaders for well-known publicly available polypharmacy side effect, interaction, and synergy prediction datasets. These datasets are listed in Table \ref{tab:datasets} with the cardinality of drug, context, and labeled triple sets. As it can be seen most of the deep drug pair scoring models are evaluated on databases that have a small number of compounds, a large number of potential combinations, and those pairings are administered in a wide range of chemical and biological contexts. The data cleaning and feature generation process are  discussed in Appendix \ref{appendix:data}.
\begin{table}[h!]
\caption{We integrated various publicly available drug pair scoring datasets in \textit{ChemicalX}. These are listed below with the domain of the pair scoring task and the number of drugs ($|\mathcal{D}|$), administration contexts ($|\mathcal{C}|$) and labeled triples $(|\mathcal{Y}|)$. }\label{tab:datasets}

\begin{tabular}{ccccc}
\hline
 \textbf{Dataset}&\textbf{Task} & $|\mathcal{D}|$  & $|\mathcal{C}|$  & $|\mathcal{Y}|$ \\
 \hline
 \textbf{TWOSIDES}    \cite{tatonetti2012data}     & Polypharmacy & 644   &  10 & 499,582 \\[0.1cm]
 \textbf{Drugbank DDI}    \cite{ryu2018deep}           & Interaction  & 1,706 &  86 & 383,496 \\[0.1cm]
 %\textbf{ZhangDDI}   \cite{zhang2017predicting}   &  &  &  &  \\[0.1cm]
 %\textbf{ChCh-Miner} \cite{biosnapnets}           &  &  &  &  \\[0.1cm]
 \textbf{DrugComb}    \cite{zagidullin2019drugcomb,zheng2021drugcomb} & Synergy&4,146  & 288 & 659,333   \\[0.1cm]
 %\textbf{SynergyXDB} \cite{seo2020synergxdb}      &  &  &  &  \\[0.1cm]
 \textbf{DrugCombDB}  \cite{liu2020drugcombdb}     & Synergy      & 2,956 & 112 & 191,391   \\[0.1cm]
 \textbf{OncolyPharm} \cite{huang2021therapeutics} & Synergy & 38 & 39 & 23,052\\[0.1cm]
 \hline
\end{tabular}

\end{table}
\subsection{Predictive Performance}\label{subsec:prediction}
In this set of experiments, our goal is to compare the architectures in the library on the three pair scoring tasks -- synergy scoring, interaction, and polypharmacology prediction. For these experiments, we are using the DrugComb, DrugbankDDI, and TWOSIDES datasets integrated in \textit{ChemicalX} -- for details see Table \ref{tab:datasets}.

\subsubsection{Experimental Settings}
The experimental details, specifically the default hyperparameters, optimizer settings, and the software package versions used in our experiments are included in Appendices \ref{appendix:software} and \ref{appendix:predictive} for reproducibility purposes. Using 80\% of the labeled instances we trained deep pair scoring models with the default settings and scored on the remainder. We computed mean predictive performances with standard errors from 10 data splits and reported the AUROC, AUPR, and F$_1$ score values in Table \ref{tab:predictive_performance}. The data splits were seeded to help the comparison of results and the F$_1$ scores used a 0.5 cutoff of propensities output by various architectures.

\subsubsection{Experimental Results}
There are multiple interesting general takeaways from the predictive performance results and the comparison presented in Table \ref{tab:predictive_performance}. These can be summarized as:

\begin{enumerate}
    \item Architectures that use pre-computed drug fingerprints as inputs for the drug encoders such as DeepSynergy \cite{preuer2018deepsynergy}, MatchMaker \cite{brahim2021matchmaker} and DeepDDI \cite{ryu2018deep} provide the best results across all of the three datasets.
    \item All of the models have good predictive performance across the tasks, even though each one of them was designed to answer specific drug pair scoring questions. For example, MatchMaker was designed for synergy scoring, yet it has a remarkably good predictive performance on the polypharmacy and interaction prediction datasets.
    \item DeepSynergy \cite{preuer2018deepsynergy} and MR-GNN \cite{xu2019mr} have the best predictive performance across all of the datasets.
\end{enumerate}

\begin{table*}[!ht!]
\centering
\caption{The predictive performance of selected deep pair scoring models in \textit{ChemicalX} on synergy scoring, interaction and polypharmacy side effect prediction tasks. We report mean predictive performances on the test set with standard errors around the mean computed from 10 seeded splits. Bold numbers denote the best performing model for each dataset and metric.}\label{tab:predictive_performance}
{\small
\begin{tabular}{l ccc ccc ccc}
        &  \multicolumn{3}{c}{\textbf{DrugComb}} & \multicolumn{3}{c}{\textbf{Drugbank DDI}} &
\multicolumn{3}{c}{\textbf{TWOSIDES}} \\
\cmidrule[0.4pt](lr{0.125em}){2-4}
\cmidrule[0.4pt](lr{0.125em}){5-7}
\cmidrule[0.4pt](lr{0.125em}){8-10}
 \textbf{Model}   & \textbf{AUROC}    & \textbf{AUPR}    & \textbf{F}$_1$  & \textbf{AUROC}    & \textbf{AUPR}    & \textbf{F}$_1$  & \textbf{AUROC}    & \textbf{AUPR}    & \textbf{F}$_1$\\ \hline

\textbf{DeepDDI} \cite{ryu2018deep}& $.669\pm .001$ & $.732\pm .001$ & $.715\pm .003$ & $.880\pm .002$ & $.837\pm .003$ & $.806\pm .002$ & $.929\pm .001$ & $.907\pm .001$ & $.848\pm .009$ \\[0.1cm]
\textbf{DeepSynergy} \cite{preuer2018deepsynergy} & $.702\pm .003$ & $\mathbf{.758\pm .003}$ & $\mathbf{.725\pm .002}$ & $\mathbf{.992\pm .001}$ & $\mathbf{.987\pm .001}$ & $\mathbf{.968\pm .001}$ & $\mathbf{.940\pm .001}$ & $\mathbf{.919\pm .001}$ & $\mathbf{.887\pm .001}$ \\[0.1cm]
\textbf{MR-GNN}\cite{xu2019mr} &$\mathbf{.744\pm .003}$ & $.574\pm .002$ & $.455\pm .002$ & $.877\pm .002$ & $.842\pm .003$ & $.821\pm .002$ & $.937\pm .002$ & $.917\pm .001$ & $.875\pm .002$ \\[0.1cm]
\textbf{SSI-DDI} \cite{nyamabossi20201ssi}& $.627\pm .001$ & $.689\pm .002$ & $.711\pm .002$ &  $.745\pm .002$ & $.723\pm .002$ & $.707\pm .003$ & $.823\pm .002$ & $.800\pm .003$ & $.756\pm .001$ \\[0.1cm]
\textbf{EPGCN-DS} \cite{sun2020structure}& $.629\pm .002$ & $.690\pm .001$ & $.697\pm .001$ & $.761\pm .002$ & $.724\pm .003$ & $.717\pm .003$& $.855\pm .003$ & $.834\pm .002$ & $.785\pm .004$  \\[0.1cm]
\textbf{DeepDrug} \cite{cao2020deepdrug}& $.643\pm .001$ & $.703\pm .002$ & $.724\pm .001$ & $.861\pm .003$ & $.827\pm .003$ & $.805\pm .002$ & $.923\pm .004$ & $.904\pm .002$ & $.857\pm .002$ \\[0.1cm]
\textbf{GCN-BMP} \cite{chen2020gcnbmp}  & $.594\pm .001$ & $.662\pm .002$ & $.707\pm .002$ & $.669\pm .001$ & $.645\pm .002$ & $.621\pm .001$ & $.709\pm .003$ & $.694\pm .002$ & $.592\pm .003$ \\[0.1cm]
\textbf{DeepDDS} \cite{wang2021deepdds}& $.663\pm .004$ & $.729\pm .002$ & $.702\pm .003$ &  $.963\pm .001$ & $.956\pm .001$ & $.910\pm .002$&$.915\pm .002$ & $.898\pm .002$ & $.839\pm .003$  \\[0.1cm]
\textbf{MatchMaker} \cite{brahim2021matchmaker} & $.662\pm .002$ & $.725\pm .001$ & $.712\pm .002$ & $.987\pm .001$ & $.981\pm .001$ & $.959\pm .001$ & $.912\pm .002$ & $.892\pm .001$ & $.849\pm .001$ \\[0.1cm]
 \hline
\end{tabular}
}
\end{table*}

\subsection{Training Runtime and Inference Scalability}
The goal of \textit{ChemicalX} is to provide fast and scalable computational tools that can learn to predict specific outcomes of drug combinations in biological and chemical contexts. We showcase the training and inference runtime of our framework using commodity hardware and discuss practical aspects of it.

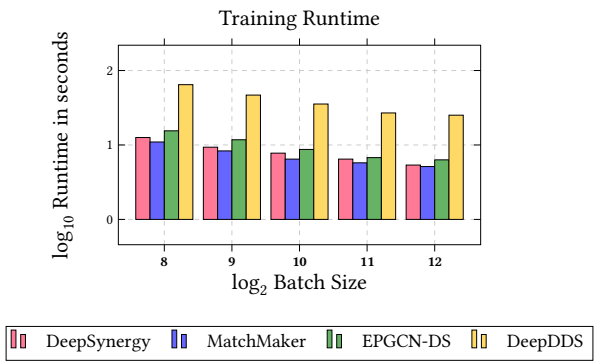
\begin{figure}[h!]

	\centering
	\begin{tikzpicture}[scale=0.6,transform shape]
	\tikzset{font={\fontsize{14pt}{12}\selectfont}}
	\begin{groupplot}[group style={group size=2 by 2,
		horizontal sep=50pt, vertical sep=70pt,ylabels at=edge left},
	width=0.54\textwidth,
	height=0.3375\textwidth,
	ymin=0,
	ymax=2,
	legend columns=4,
every tick label/.append style={font=\bf},
    y tick label style={
        /pgf/number format/.cd,
            fixed,
            fixed zerofill,
            precision=0,
        /tikz/.cd
    },
 enlarge x limits=true,
	grid=major,
	grid style={dashed, gray!40},
	scaled ticks=false,
	inner axis line style={-stealth}]

 \nextgroupplot[
   xlabel=$\log_2$ Batch Size,
    ybar=0pt,
      every tick/.style={
        black,
        semithick,
      },
    bar width=9pt,
    enlargelimits=0.17,
    legend columns=4,
    legend image post style={solid},
    legend style={at={(0.5,-0.25)},nodes={scale=1.5, transform shape}, 
      anchor=north,legend columns=-1},
    ylabel={$\log_{10}$ Runtime in seconds},
ytick={0,1,2,3},
title=Training Runtime,
    symbolic x coords={8, 9, 10, 11, 12},
    xtick={8, 9, 10, 11, 12},
    	legend style = { column sep = 10pt, legend columns = 1, legend to name = grouplegend, font=\small}  ]

\addplot [fill=awesome!60]  coordinates {
(8,1.10)
(9,0.97)
(10,0.89)
(11,0.81)
(12,0.73)
};
\addlegendentry{DeepSynergy}
\addplot [fill=blue!60]  coordinates {
(8,1.04)
(9,0.92)
(10,0.81)
(11,0.76)
(12,0.71)

};
\addlegendentry{MatchMaker}
\addplot [fill=ao(english)!60] coordinates {
(8,1.19)
(9,1.07)
(10,0.94)
(11,0.83)
(12,0.80)
};\addlegendentry{EPGCN-DS}
\addplot [fill=amber!60] coordinates {

(8,1.81)
(9,1.67)
(10,1.55)
(11,1.43)
(12,1.40)};
\addlegendentry{DeepDDS}

	\end{groupplot}

	\node at ($(group c1r1) + (0cm,-4.4cm)$) {\ref{grouplegend}}; 
	\end{tikzpicture}
	
	\caption{The average runtime of doing a whole epoch on a sample of the DrugBankDDI dataset as a function of the batch size - the average runtime values were computed from 10 experimental runs for each of the models. }\label{fig:runtime}
\end{figure}

\subsubsection{Experimental Settings -- Training Runtime.} Using $2^{17}$ randomly selected drug pairs from DrugBankDDI \cite{ryu2018deep} we do a single epoch using the DeepSynergy \cite{preuer2018deepsynergy}, MatchMaker \cite{brahim2021matchmaker}, EPGCN-DS \cite{sun2020structure}, and DeepDDS \cite{wang2021deepdds} models with varying batch sizes. The hyperparameter and optimizer settings were taken from Appendix \ref{appendix:predictive}. The mean runtime calculated from 10 experimental runs on a CPU is plotted on Figure \ref{fig:runtime} as a function of the batch size for each of the models. Details about the software package versions and hardware used in the  experiments are discussed in Appendices \ref{appendix:software} and \ref{appendix:hardware}.

\subsubsection{Experimental Results -- Training Runtime.} Our results in Figure \ref{fig:runtime}  demonstrate that the various drug pair scoring models implemented in \textit{ChemicalX} can be trained on combination datasets with millions of drug pairs in a few minutes using commodity hardware. Moreover, these findings support that using larger batches of drug pairs can reduce the training runtime moderately. There is strong evidence that the graph convolutional models such as EPGCN-DS and DeppDDS suffer from longer runtimes. This finding combined with the previously observed low predictive performance strongly suggests that traditional architectures such as DeepSynergy and MatchMaker are superior on both the predictive and computational performance aspects for practical use cases.

\subsubsection{Experimental Settings -- Inference Scalability.} 
\begin{figure}[h!]
\centering
\scalebox{0.85}{
\begin{tikzpicture}
\begin{groupplot}[	grid=major,
	grid style={dashed, gray!40},group style={
                      group name=myplot,
                      group size= 1 by 1, horizontal sep=1.55cm,vertical sep=1.2cm},height=4.2cm,width=6.3cm, title style={at={(0.5,1.0)},anchor=south},every axis x label/.style={at={(axis description cs:0.5,-0.15)},anchor=north},]
\nextgroupplot[
	ylabel=Runtime in hours,
	xlabel=$\log_2$ Number of drugs,
	xtick={9,10,11,12},
	xmin=8.8,
	xmax=12.2,
	title=Inference Runtime,
	ymin=-0.2,
	ymax=2.2,
	ytick={0,0.5,1,1.5,2},
 	legend columns=4,
	legend style={at={(0.50,-0.6)},anchor=south},
y label style={at={(0.05,0.5)}},
    legend entries={$\text{DeepSynergy}$, $\text{MatchMaker}$, $\text{EPGCN-DS}$,$\text{DeepDDS}$},
]
\addplot [very thick, awesome,mark=*,opacity=0.6]coordinates {
(9.0,0.005)
(9.333,0.007)
(9.667,0.012)
(10.0,0.019)
(10.333,0.03)
(10.667,0.047)
(11.0,0.075)
(11.333,0.119)
(11.667,0.188)
(12.0,0.299)
};
\addplot [very thick, blue,mark=square*,opacity=0.6]coordinates {
(9.0,0.005)
(9.333,0.009)
(9.667,0.014)
(10.0,0.022)
(10.333,0.035)
(10.667,0.055)
(11.0,0.087)
(11.333,0.139)
(11.667,0.22)
(12.0,0.35)
};
\addplot [very thick, ao(english), ,mark=triangle*,opacity=0.6]coordinates {
(9.0,0.009)
(9.333,0.015)
(9.667,0.023)
(10.0,0.037)
(10.333,0.059)
(10.667,0.093)
(11.0,0.148)
(11.333,0.234)
(11.667,0.372)
(12.0,0.59)
};
\addplot [very thick, amber,mark=diamond*,opacity=0.6]coordinates {
(9.0,0.028)
(9.333,0.045)
(9.667,0.071)
(10.0,0.113)
(10.333,0.180)
(10.667,0.186)
(11.0,0.454)
(11.333,0.721)
(11.667,1.144)
(12.0,1.815)
};

\end{groupplot}
\end{tikzpicture}}
\caption{The average estimated runtime of doing a whole scoring pass for all of the possible drug pair combinations in a drug set for a single context as a function of the drug set cardinality.}\label{fig:times}
\end{figure}
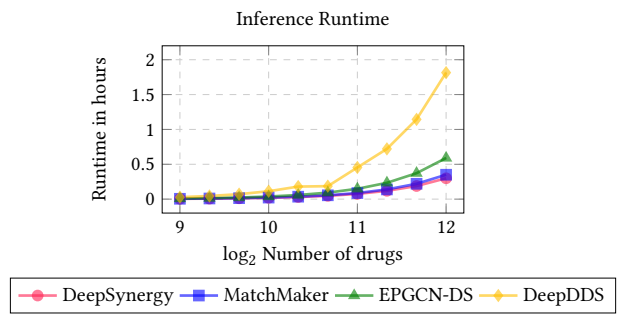

Using the experimental settings from runtime measurements we also estimated the average time needed to do inference on a labeled triple set of $2^{17}$ drug pairs. Given this estimate, a set of drugs, and a single administration context we  computed the time needed to perform inference on all of the drug pairs in this given drug set as a function of the cardinality of the drug set. We plotted the estimated inference times in hours against the number of drugs in the drug set on Figure \ref{fig:times} for the DeepSynergy \cite{preuer2018deepsynergy}, MatchMaker \cite{brahim2021matchmaker}, EPGCN-DS \cite{sun2020structure} and DeepDDS \cite{wang2021deepdds} models. We assume that scoring happens in drug pair batches of size $2^{12}$ and that the number of possible drug pairs given a drug set is the number of drugs squared.

\subsubsection{Experimental Results -- Inference Scalability.} The runtime of inference scales quadratically with the number of drugs as Figure \ref{fig:times} shows; this is natural as the number of pair combinations also scales quadratically with the number of drugs in the drug set. Our results imply that architectures with feedforward neural network-based drug encoders could score all of the pairs which can be formed by combining Federal Drug Administration approved drugs in a matter of half an hour, while graph convolutional models could achieve the same in a few hours. It is worth emphasizing that given a trained model the drug pair scoring inference can be trivially distributed and run in parallel which means that even graph convolutional models can score all possible pairs from a drug set of $2^{12}$ drugs in a few minutes on larger industry-scale computing clusters.

\section{Impact and Future Directions}\label{sec:future}
The streamlined design and open-source nature of the framework open up new possibilities for drug discovery practitioners and create opportunities for future research into the domain.

\subsection{Potential users}
When we designed \textit{ChemicalX} we had specific user stories in mind inspired by our daily work in AstraZeneca. Here, we discuss potential applications of the deep learning framework in various early phases of the drug discovery process.
\subsubsection{Oncologists.} A range of models available in \textit{ChemicalX} was designed to solve the drug combination synergy prediction problem in oncology. In this setting, a pair of drugs is administered to destroy a cancer cell and the related machine learning task is to predict the synergistic nature of the relationship. Given the prohibitive number of drug pair-context triples, the \textit{in vitro} evaluation of combinations is impossible. However, an \textit{in silico} approach which provides indications that are learned from data can reduce the search space radically.
\subsubsection{Computational Chemists.} The prediction of potential drug-drug interactions could be particularly useful in the lead optimization phase of the drug development process. Computational chemists can exploit the indications output by \textit{ChemicalX} models to flag potential unwanted drug-drug interactions in advance before the drug enters the pre-clinical phase. Taking into account the high attrition rate of drug development programs after the lead optimization phase \cite{gaudelet2021utilizing}, early warnings by computational systems can mitigate risks.

\subsubsection{Drug Safety Researchers.} The identification of rare unexpected polypharmacy side effects is not part of the drug development process. This is mainly due to a large number of potential drug pair-side effect combinations which need to be investigated. Because of this, those computational methods in \textit{ChemicalX} that find potential indications of polypharmacy-related adverse events are highly valuable for drug safety researchers. 

\subsection{Ensemble models}
The deep learning models in \textit{ChemicalX} exploit various sources of information -- molecule level features, graph structure, and local patterns. Moreover, our results in Section~\ref{sec:experiments} have demonstrated that models have significantly different performances on task and across tasks. It is reasonable to assume that the predictive performance of models is dependent on the type of molecules. Hence, ensembling the techniques to potentially increase the predictive performance on the pair scoring task would be an important future direction for machine learning research.
\subsection{Set-based generalization}
Our framework focuses on predicting outcomes for the simplest type of drug combinations - pairs. However, drugs are commonly administered in numbers greater than two, hence, investigating context-specific outcomes is highly relevant in these situations. Some of the models in our framework such as \textit{DeepSynergy} \cite{preuer2018deepsynergy} could be generalized to handle sets of drugs as input. It is worth noting, however, that there are not many large-scale higher-order drug combination datasets \cite{liu2020drugcombdb} that are readily available in the public domain. 
\subsection{Transferring architectures between tasks}
The discussion about the unified model of drug pair scoring highlighted that pair scoring is done by the \textit{head layer} and the goal of the \textit{drug encoder} is the generation of drug features. Earlier results about pre-training molecular representation models \cite{hu2019strategies} demonstrated that using a data-rich domain for fitting the encoder and fine-tuning on another task can lead to state-of-the-art results. We hypothesize that the same way \textit{ChemicalX} opens up possibilities for transferring knowledge between drug pair scoring tasks - the encoder is trained on one task and the scoring head is fine-tuned on another one.

\section{Conclusions}\label{sec:conclusions}
In this paper, we discussed \textit{\textit{ChemicalX}}, the first deep learning library dedicated to solving the drug pair scoring task. We gave an overview of related literature on geometric deep learning, drug pair scoring, and machine learning frameworks; we also formalized the drug pair scoring task itself. We highlighted the general design principles of \textit{ChemicalX}; the integrated benchmark datasets, custom data structures, neural network layers, and model designs. We discussed how test-driven development, continuous integration, detailed documentation, and using modern software engineering practices promote the long-term viability of the project. Our experimental evaluation of \textit{ChemicalX} concentrated on: (a) the predictive performance of drug pair scoring models on drug-drug interaction, polypharmacy side effect, and synergy prediction tasks; (b) the run time of these models as a function of drug pair database sizes.

\begin{acks}
The authors would like to thank Kexin Huang and Zhaocheng Zhu for help and feedback throughout the preparation of this manuscript. Charles Tapley Hoyt, Klas Karis, and Benjamin M. Gyori were supported by the DARPA Young Faculty Award W911NF20102551.

\end{acks}

\clearpage 

\bibliographystyle{ACM-Reference-Format}

\bibliography{main}

\appendix

\section{Data Integration}\label{appendix:data}
The Python scripts used for data cleaning and integration are available in the package repository to help the with reproducibility.

\subsection{Drug feature generation}
For each of the drugs in the labeled triple sets we retrieved the canonical SMILES strings \cite{weininger1988smiles}. Labeled triples for which this molecular representation was not available were discarded. We computed $2^8$ dimensional hashed Morgan fingerprints for each drug which used a radius of 2. Finally, using the SMILES strings of the compounds we generated molecular graphs using TorchDrug.

\subsection{Negative samples}
The DrugBankDDI and TwoSides datasets do not contain ground truth negative samples, for these we generated negative samples without collisions that equal the number of triples with positive labels. Negative samples are generated uniformly and discarded if collisions happened with the existing ground truth labels.

\section{Software Package Versions}\label{appendix:software}
The predictive performance and runtime experiments in Section \ref{sec:experiments} were done with the 0.1.0 \textit{ChemicalX} release. We specifically used the following  Python package versions listed in Table \ref{tab:package}.
\begin{table}[h!]

\caption{The Python package versions used in the experimental evaluation of ChemicalX.}\label{tab:package}

\begin{tabular}{cc}
\hline
\textbf{Package name} & \textbf{Package version} \\
\hline 
   Numpy     &   1.19.2 \\
    Pandas    &   1.3.0      \\
    Torch    &     1.10.1    \\
    Torch Scatter    & 2.0.9  \\
    Torch Drug & 0.1.2 \\
    \hline 
\end{tabular}
\end{table}

\section{Predictive Performance Experimental Details}\label{appendix:predictive}
In Subsection \ref{subsec:prediction} we trained all of the pair scoring models in \textit{ChemicalX} using the same dropout rate \cite{srivastava2014dropout}, settings of the Adam optimizer \cite{kingma2015adam}, batch size and number of epochs. These hyperparameter values are listed in Table \ref{tab:hyperparameters}. 

\begin{table}[h!]
\caption{The default model training settings used in the predictive performance evaluation of pair scoring models implemented in \textit{ChemicalX}.}\label{tab:hyperparameters}
\begin{tabular}{lc}

           & \textbf{Default setting}   \\
\hline
\textbf{Epochs}             & $50$    \\
\textbf{Batch size}         & $2^{12}$     \\
\textbf{Dropout}            & 0.5       \\
\textbf{Learning rate}      & $10^{-2}$ \\
\textbf{Weight decay}       & $10^{-5}$            \\
\textbf{Adam optimizer} $\beta_1$     & $0.9$               \\
\textbf{Adam optimizer} $\beta_2$     & $0.99$            \\
\textbf{Adam optimizer} $\varepsilon$ & $10^{-7}$               \\
\hline
\end{tabular}
\end{table}

The deep pair scoring models implemented in the 0.1.0 \textit{ChemicalX} release have default hyperparameter settings. We used these settings in Section \ref{subsec:prediction} when we investigated the experimental evaluation of the predictive performance -- these are listed in Table \ref{tab:modelhyperparams}. In each of the models we used rectified linear unit activations \cite{nair2010rectified} in the encoder and intermediate head layers.
\begin{table}[h!]
\caption{The default hyperparameters of deep pair scoring models implemented in \textit{ChemicalX}. We used these settings in the predictive performance and scalability evaluation experiments.}\label{tab:modelhyperparams}
\begin{tabular}{ccc}
\hline
\textbf{Model}                        & \textbf{Hyperparameter}           & \textbf{Value}        \\
\hline
\textbf{DeepDDI} \cite{ryu2018deep}   &            Hidden layer channels              &    $\left\{2^5,2^5,2^5,2^5\right\}$          \\[0.3em]
\hline
\multirow{3}{*}{\textbf{DeepSynergy} \cite{preuer2018deepsynergy}} & Drug encoder channels    & $2^7$          \\[0.3em]
                             & Context encoder channels & $2^7$          \\[0.3em]
                             & Hidden layer channels    & $\left\{2^5,2^5,2^5\right\}$ \\[0.3em]
                             \hline
                        
            \multirow{4}{*}{\textbf{MHCADDI}      \cite{deac2019mhcaddi} }       
            
        &Atom encoder channels&  $2^4$\\[0.3em]
        &Edge encoder channels & $2^4$\\[0.3em]
        &Hidden layer channels&  $2^4$\\[0.3em]
        &Readout layer channels & $2^4$\\[0.3em]
                             \hline
         \multirow{3}{*}{ \textbf{MRGNN}         \cite{xu2019mr}      }            &Drug encoder channels&  $2^5$\\[0.3em]
        &Drug encoder layers& $2^2$\\[0.3em]
        &Hidden layer channels&  $2^4$\\[0.3em] \hline

         \multirow{4}{*}{\textbf{CASTER}\cite{huang2019caster} }                                         & Drug encoder channels & $\left\{2^5,2^5\right\}$           \\[0.3em]
                             & Hidden layer channels    & $\left\{2^5,2^5\right\}$ \\[0.3em]
                 & Regularization coefficient& $10^{-5}$           \\[0.3em]
                             & Magnification factor   & $10^2$ \\[0.3em]
                             \hline  
         \multirow{2}{*}{\textbf{SSI-DDI}      \cite{nyamabossi20201ssi}}               & Drug encoder channels & $\left\{2^5,2^5\right\}$           \\[0.3em]
                             & Attention heads   & $\left\{2,2\right\}$ \\[0.3em]
                             \hline   
\multirow{2}{*}{\textbf{EPGCN-DS} \cite{sun2020structure}}  &         Drug encoder channels                 &   $2^7$           \\[0.3em]
                             &Hidden layer channels                          &  $\left\{2^5,2^5\right\}$            \\[0.3em]\hline
\multirow{2}{*}{\textbf{DeepDrug}  \cite{cao2020deepdrug} }       &         Drug encoder channels                 &  $\left\{2^5,2^5,2^5,2^5\right\}$         \\[0.3em]
                             &Hidden layer channels                          &  $2^6$            \\[0.3em]\hline
\multirow{2}{*}{\textbf{GCN-BMP} \cite{chen2020gcnbmp}}  &         Drug encoder channels                 &   $2^4$           \\[0.3em]
                             &Hidden layer channels                          &  $2^4$            \\[0.3em]
                             \hline
\multirow{2}{*}{\textbf{DeepDDS} \cite{wang2021deepdds}}   
                             & Context encoder channels & $\left\{2^9,2^8,2^7\right\}$           \\[0.3em]
                             & Hidden layer channels    & $\left\{2^9,2^7\right\}$ \\[0.3em]
                             \hline
\multirow{2}{*}{\textbf{MatchMaker} \cite{brahim2021matchmaker}}  &         Drug encoder channels                 &   $\left\{2^5,2^5\right\}$           \\[0.3em]
                             &Hidden layer channels                          &  $\left\{2^6,2^5\right\}$            \\[0.3em]\hline
\end{tabular}
\end{table}

\section{Experimental Hardware}\label{appendix:hardware}
All of the predictive performance, training and inference runtime experiments were executed on a MacBook Pro with 2.3 GHz 8-Core Intel Core i9 processors and 16 GB 2667 MHz DDR4 memory without GPU acceleration.

\end{document}